\documentclass[conference]{IEEEtran}
\IEEEoverridecommandlockouts
\usepackage{cite}
\usepackage{amsmath,amssymb,amsfonts}
\usepackage{algorithmic}
\usepackage{graphicx}
\usepackage{textcomp}
\usepackage{xcolor}
\usepackage[hidelinks]{hyperref}
\def\BibTeX{{\rm B\kern-.05em{\sc i\kern-.025em b}\kern-.08em
    T\kern-.1667em\lower.7ex\hbox{E}\kern-.125emX}}
\begin{document}

\title{Hybrid Unet-Transformer Model for Generating Stress and Strain Fields from Composite Geometrics\\
}

\author{
\IEEEauthorblockN{ Shrey Patel}
\IEEEauthorblockA{
\textit{University of Maryland-College Park}\\
United States of America \\
ORCID: 0009-0003-0886-4583}}

\maketitle

\begin{abstract}
Accurate prediction of stress and strain fields in hierarchical 
composite microstructures is critical for physics-informed material 
design, yet conventional finite element method (FEM) simulations 
are computationally prohibitive at scale, requiring minutes to days 
per evaluation. In this work, we propose a hybrid UNet-Transformer 
architecture that predicts complex mechanical field distributions 
directly from composite microstructure geometry images, serving as 
an efficient surrogate for FEM across ten distinct stress and strain 
field types spanning diverse two-phase composite configurations 
including square, hexagonal, and triangular tessellations, multiple 
boundary conditions, and high-resolution geometries. Results demonstrate 
that the proposed architecture achieves strong predictive performance 
across the majority of subdatasets, with peak accuracy on periodic 
tessellation geometries reaching R² = 0.9991, SSIM = 0.9936, and 
MAE = 0.0050 on the boundary condition subdataset and the triangular 
tessellation subdataset respectively. Across six of the eight 
evaluated subdatasets, MAE remains below 0.05 on the normalized 
[0, 1] pixel scale. Encoder attention analysis via Grad-CAM and 
Grad-CAM++ confirms that the model develops physically meaningful 
internal representations, localizing attention at mechanically 
critical regions  including phase boundaries, ligament junctions, 
and indenter contact zones  without explicit structural supervision. 
Performance degrades on irregular square-grid geometries with sparse 
soft-phase inclusions, with the S11 normal stress subdataset yielding 
R² = 0.7735 and SSIM = 0.7126, consistent with the known limitation 
of smooth-loss image translation models in reproducing sharp stress 
discontinuities. These results establish hybrid UNet-Transformer 
models as a viable deep learning surrogate for mechanical field 
prediction in composite materials, with inference time reduced from 
minutes-to-days per FEM simulation to under one second per prediction.
\end{abstract}


\section{Introduction}
Composite materials are non-homogeneous mixtures of metals, ceramics or polymers. The combination of any constituent materials into a single composite material can significantly enhance mechanical performance. By spatially distributing soft and brittle constituent phases, composites can be engineered to achieve properties such as high stiffness, recoverability under compressive loading and crack resistance. Various material designs, such as modern ceramics, fiber-reinforced polymers and biologically evolved materials, are driven by this tunability. The advent of additive manufacturing has further expanded the accessible design space by enabling fabrication of composites with complex, hierarchical microstructures that were previously unattainable through conventional manufacturing techniques. However, as the design space grows in complexity and dimensionality, the challenge of identifying optimal microstructural configurations becomes increasingly difficult without efficient computational tools.

Realizing the full potential of composite design requires accurate knowledge of how stress and strain distribute spatially across a heterogeneous microstructure under mechanical loading. Failure initiation, recoverability and local stress concentration are all governed by physical field distributions, and thus require a physics-informed design process. Finite Element Method (FEM) simulations are the standard tool for computing these fields, providing high-fidelity ground truth for arbitrary geometries and loading conditions. However, a single FEM simulation for one composite geometry is already computationally expensive, and design optimization or uncertainty quantification typically requires evaluating thousands of candidate configurations. The resulting computational cost is combinatorially large, making exhaustive FEM evaluation of candidate designs computationally intractable. These challenges have strongly motivated the development of deep learning architectures as efficient surrogates for FEM simulations, capable of predicting physical field distributions at a fraction of the computational cost once trained.

Recent advances in deep learning have fundamentally transformed surrogate modeling across computational science and engineering. Deep learning models have demonstrated remarkable success in approximating complex physical phenomena across domains including fluid dynamics, molecular dynamics, structural mechanics and materials science, consistently achieving accuracy comparable to high-fidelity numerical solvers at a fraction of the computational cost. The ability of deep neural networks to learn highly nonlinear mappings from large datasets of simulation outputs has established them as powerful tools for accelerating physics-based design workflows. Image-to-image translation architectures in particular have proven effective at learning spatial mappings between input configurations and output physical fields, enabling direct prediction of distributed quantities such as temperature, pressure, velocity and stress from geometric or boundary condition inputs. These capabilities have made deep learning an attractive framework for replacing or augmenting computationally expensive numerical simulations across a broad range of engineering applications.

Vision Transformers and hybrid Unet-Transformer architectures, which apply self-attention mechanisms across spatial patch sequences, have demonstrated strong performance in dense prediction tasks in computer vision such as semantic segmentation and depth estimation. However, their application to mechanical field prediction in composite materials remains largely unexplored. In this work, a hybrid Unet-Transformer is applied to the problem of stress and strain field prediction in hierarchical composites. The encoder uses residual convolutional blocks to extract multi-scale spatial features, a bottleneck transformer module with sinusoidal 2D positional embeddings captures global spatial relationships, and a convolutional decoder with skip connections and multi-scale supervision reconstructs the output field at full resolution. The dataset has been obtained from Yang et al.[1]  where a conditional GAN model was trained for predicting stress and strain fields across a range of two-phase composite geometries. The primary contribution of this work is an empirical demonstration that a hybrid Unet-Transformer achieves competitive field prediction accuracy on this benchmark, while offering a simpler, single-network training procedure.

\section{Literature Review}
The prediction of stress and strain fields in composite materials 
from microstructural geometry has emerged as a key research direction 
in data-driven mechanics, driven by the prohibitive computational 
cost of repeated finite element method (FEM) simulations [7] 
across large composite design spaces. Early work demonstrated that 
convolutional neural networks could serve as efficient surrogates 
for FEM at high accuracy. A foundational benchmark was established 
in [6], proposing SCSNet and StressNet  two CNN architectures 
for predicting von Mises stress fields in 2D linear elastic 
cantilevered structures from geometry, loading, and boundary condition 
inputs  with a mean relative error of 2.04\% on the test set, 
demonstrating that image-based deep learning could approximate FEM 
outputs for structured geometries without iterative numerical solving.

Subsequent work extended this capability to composite microstructures, 
where the heterogeneous spatial arrangement of constituent phases 
introduces substantially greater field complexity. A U-Net 
encoder-decoder architecture was proposed in [5, 15] for von Mises 
stress field prediction in fiber-reinforced composites, demonstrating 
cross-complexity generalization to higher fiber-count configurations 
unseen during training. Critically, this work established that data 
augmentation in mechanics field prediction must be restricted to 
physics-consistent transformations: axis-aligned flipping operations 
that preserve mechanical boundary conditions are permissible, whereas 
arbitrary rotation decouples the geometry image from its physically 
valid field output and constitutes a physics-violating 
transformation [15].

The image-to-image translation paradigm for composite field prediction 
was advanced in [1], where a conditional GAN comprising a U-Net 
generator and PatchGAN discriminator was trained to predict von Mises 
stress and plastic strain fields in two-phase hierarchical composites 
directly from 2D geometry images, using FEM simulations in Abaqus 
as ground truth across 2,000 randomly generated composite geometries. 
Generalization was demonstrated to unseen geometries, multiple loading 
conditions, and non-square tessellations including hexagonal and 
triangular unit arrangements. A follow-up study [2] extended this 
to predict complete stress and strain tensor components across all 
field directions, incorporating data statistics tuning to improve 
generalization across varying constituent ratios under limited 
training data.

Alternative architectural paradigms for the same geometry-to-field 
mapping task were explored in parallel. A Fourier Neural Operator 
framework was proposed in [3] that operates in the frequency domain 
rather than the pixel domain, achieving high-fidelity predictions 
of complete stress and strain fields with very few training samples 
and demonstrating zero-shot generalization to unseen geometries and 
zero-shot super-resolution  capabilities not accessible to 
pixel-domain convolutional surrogates. A graph neural network 
framework was proposed in [4] that exploits the natural 
correspondence between FEM meshes and graphs, predicting deformation, 
stress and strain fields across fiber composites, stratified 
composites and lattice metamaterials, and capturing nonlinear 
phenomena including plasticity and buckling instability.

Despite this progression, a common limitation across the above 
architectures is their reliance on convolutional inductive biases  
local receptive fields and translational equivariance  which 
constrain the model's capacity to capture long-range spatial 
dependencies in stress and strain fields, particularly for geometries 
with non-local load transfer paths. Vision Transformers address 
this through global self-attention mechanisms with no locality 
constraint, making them in principle better suited to capturing 
long-range mechanical interactions. However, standard Vision 
Transformers require large training datasets to converge effectively, 
which is incompatible with the small FEM-generated datasets typical 
of computational mechanics. A hybrid CNN-Transformer architecture 
for small data paradigms was proposed in [10], incorporating 
progressive tokenization with feature supplementation and a 
convolutional prediction module shortcut, enabling effective 
training without large-scale pre-training.

The present work builds directly on this progression. The dataset 
from [1] gives 22,000 FEM-generated composite geometry and field image 
pairs across ten subdatasets  is adopted, and the cGAN 
image-to-image architecture is replaced with a hybrid UNet-Transformer, 
providing the first direct architectural comparison between 
convolutional image translation and global-attention-based field 
prediction on this benchmark. Prediction quality is evaluated 
using MAE, SSIM [8], and $R^2$, and encoder attention is 
interpreted using Grad-CAM [9] and Grad-CAM++ [11] to assess 
whether global attention produces physically meaningful spatial 
localization without explicit structural supervision.

\section{Proposed Methodology}
The proposed model,  the hybrid Unet-Transformer, takes a 128×128 RGB composite geometry image as input and produces a 128×128 RGB stress or strain field image as output, framing mechanical field prediction as an image-to-image translation problem. The training ran for around 4 hours 25 minutes on Google Colab's T4 GPU. Code availability : "https://anonymous.4open.science/r/hybrid-Unet-Transformer-Stress-Field-Prediction-C112/". Data is uploaded on huggingface as well which the notebook in the paper directly accesses it.

\subsection{Data Collection and Preprocessing}\label{AA}
The dataset used in this work is sourced from Yang et al. [1], where it was generated through finite element method (FEM) simulations using the commercial Abaqus software. The composite geometries consist of two constituent phases  brittle units and soft units  arranged on a two-dimensional grid in the x-y plane. The brittle units are represented in white and the soft units in red against a black background, providing unambiguous encoding of the microstructural geometry as an RGB image. Out of 2³² possible geometric combinations arising from the binary arrangement of soft and brittle units on the grid, 2,000 geometries were randomly selected to form the dataset. The ratio of soft units across the dataset follows an approximately Gaussian distribution, centered around 0.5, ensuring a balanced representation of microstructural configurations across the design space.

\begin{figure}[htbp]
\centerline{\includegraphics[width=0.45\textwidth]{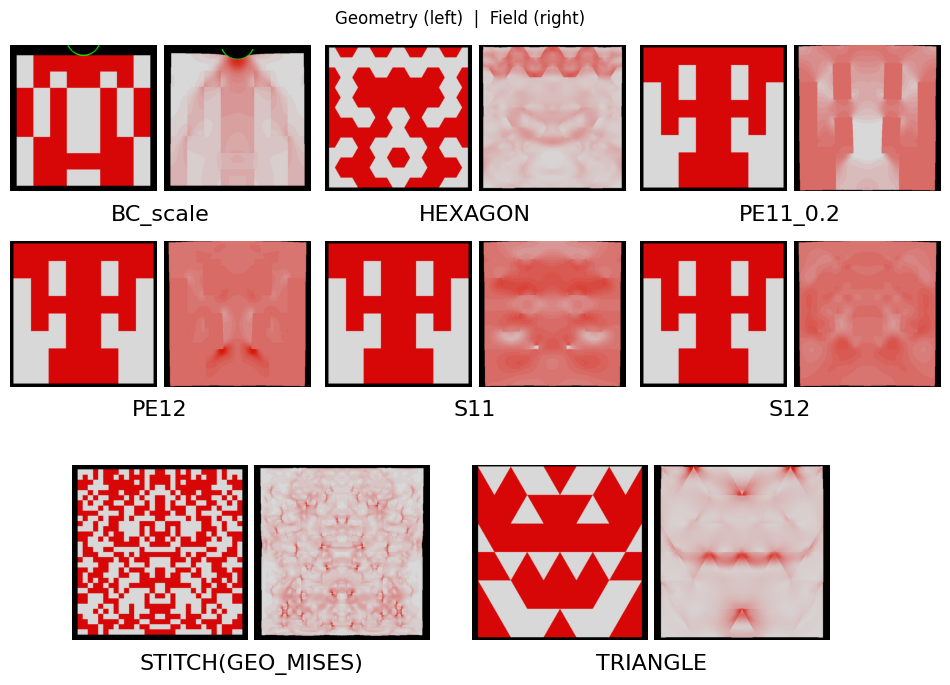}}
\caption{(a) Geometry on the left half (b) FEM Field on right half}
\label{fig:dataset}
\end{figure}

The dataset covers eight distinct field types, each constituting a separate subdataset.The PE11 subdataset captures the plastic strain component in the x-direction, encoding directional deformation along the horizontal loading axis. PE12 provides the shear strain component, reflecting off-axis deformation between the x and y directions. S11 contains the normal stress in the x-direction, representing the direct stress component along the compressive loading axis, while S12 captures the corresponding shear stress between the x and y directions. Beyond these field-type variants, the dataset includes three geometry-variant subdatasets. The hexagonal subdataset contains composites built from hexagonal tessellations in place of square units, testing the model's generalization to non-square elementary shapes. The triangular subdataset similarly replaces square units with triangular tessellations. The boundary condition subdataset encodes multiple loading scenarios  specifically compression and nanoindentation  directly within the geometry image using green lines as rigid body indicators, enabling a single model to distinguish and respond to distinct mechanical boundary conditions. The von Mises subdataset uses a higher-resolution 32×32 grid representation in place of the base 8×8 grid, allowing the model to capture more complex microstructural patterns. All subdatasets share a common storage format: each image is stored as a side-by-side PNG, where the left half encodes the input composite geometry and the right half encodes the corresponding FEM-computed stress or strain field, as shown in Figure ~\ref{fig:dataset}.

No physics-altering preprocessing or artificial augmentation techniques such as rotation, colour jitter or noise injection are applied, as such transformations would decouple the geometry image from its physically consistent field output and cause the model to learn spurious mappings inconsistent with the underlying mechanics [15].

\subsection{Model Architecture and Training}
A hybrid Unet-Transformer architecture is proposed, designed specifically for image-to-image translation from FEM geometry inputs to stress/strain field outputs[13]. The model accepts a 3-channel 128×128 geometry image and produces a corresponding 3-channel 128×128 field prediction. The architecture consists of three components: a convolutional encoder, a Transformer bottleneck, and a multi-scale convolutional decoder with skip connections. Due to limited data of only 22,000 images in total, a hybrid architecture likes this provides better performance over Vision transformer [10]. Here, the Unet[12] encoders will capture the local features, whereas transformer block will attend to global context of FEM geometric images.

\begin{figure}[htbp]
\centerline{\includegraphics[width=0.45\textwidth]{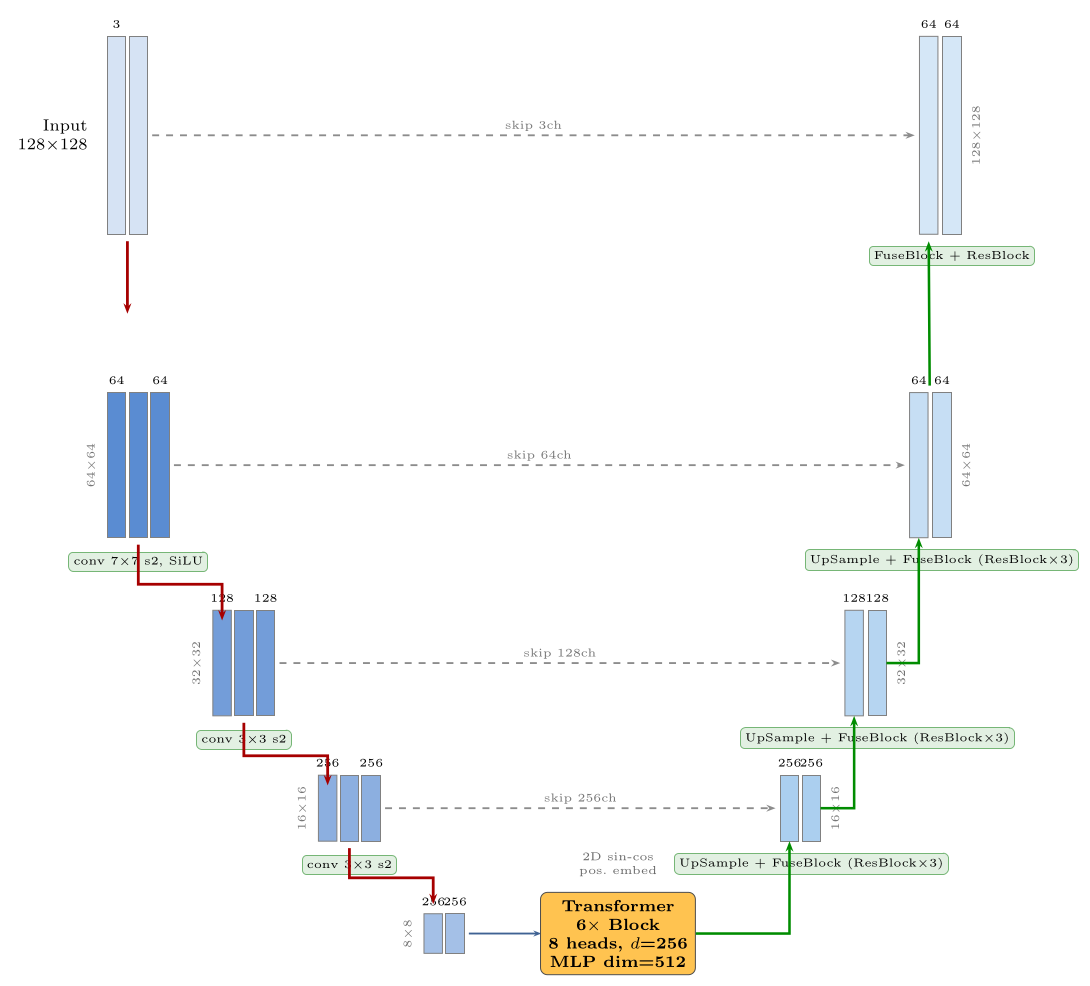}}
\caption{Model Architecture}
\label{fig:Model}
\end{figure}

\subsubsection{Encoder}
The encoder progressively downsamples the input from 128×128 to an 8×8 spatial bottleneck through four stages, producing skip features at each intermediate resolution as shown in Figure ~\ref{fig:Model}. Each residual blocks consists of two convolution layers interleaved with GroupNorm and SilU activations, with a residual shortcut connections.
Each ResBlock computes:
\begin{equation}
    \mathbf{x}' = \mathbf{x} + \mathrm{GN}_2\!\left(\mathrm{Conv}_{3\times3}\!\left(\mathrm{SiLU}\!\left(\mathrm{GN}_1\!\left(\mathrm{Conv}_{3\times3}(\mathbf{x})\right)\right)\right)\right)
\end{equation}
where $\mathrm{GN}(\cdot)$ denotes Group Normalization with 8 groups,
$\mathrm{SiLU}(\cdot) = x \cdot \sigma(x)$ is the Sigmoid Linear Unit
activation, and $\mathbf{x}$ is the identity shortcut.

\subsubsection{Transformer Bottleneck}
The 8×8 feature map from the encoder is flattened into a sequence of 64 tokens of dimension 256 and augmented with fixed 2D sinusoidal positional embeddings. The positional encoding follows the standard formulation using sine and cosine functions of frequencies spaced logarithmically between 1 and 10,000, computed independently for both spatial dimensions (y and x) and concatenated.

Six TransformerBlock modules are applied sequentially. Each block uses pre-norm LayerNorm, multi-head self-attention, and a two-layer MLP with GELU activation (hidden dimension 512). Residual connections are applied around both the attention and MLP sub-layers. A learnable linear projection is additionally applied to the positional-embedded tokens before the Transformer layers.

Given a sequence of tokens $\mathbf{Z} \in \mathbb{R}^{N \times d}$ where
$N = H \cdot W = 64$ and $d = 256$, each TransformerBlock computes:
\begin{align}
    \mathbf{Z}' &= \mathbf{Z} + \mathrm{MHSA}\!\left(\mathrm{LN}(\mathbf{Z})\right) \\
    \mathbf{Z}'' &= \mathbf{Z}' + \mathrm{MLP}\!\left(\mathrm{LN}(\mathbf{Z}')\right)
\end{align}
where $\mathrm{LN}(\cdot)$ is Layer Normalization (pre-norm), and
$\mathrm{MHSA}(\cdot)$ is Multi-Head Self-Attention with $h = 8$ heads defined as:
\begin{equation}
    \mathrm{MHSA}(\mathbf{Z}) = \mathrm{Concat}(\mathrm{head}_1, \ldots, \mathrm{head}_h)\,\mathbf{W}^O
\end{equation}
\begin{equation}
    \mathrm{head}_i = \mathrm{softmax}\!\left(\frac{\mathbf{Q}_i \mathbf{K}_i^\top}{\sqrt{d_k}}\right)\mathbf{V}_i
\end{equation}
where $d_k = d / h = 32$ is the per-head dimension, and $\mathbf{W}^O \in \mathbb{R}^{d \times d}$ are learned projection matrices.

The MLP sub-layer is a two-layer feed-forward network:
\begin{equation}
    \mathrm{MLP}(\mathbf{z}) = \mathbf{W}_2\,\mathrm{GELU}(\mathbf{W}_1 \mathbf{z} + \mathbf{b}_1) + \mathbf{b}_2
\end{equation}
with hidden dimension $d_{\mathrm{mlp}} = 512$
(i.e., $\mathbf{W}_1 \in \mathbb{R}^{512 \times 256}$,
$\mathbf{W}_2 \in \mathbb{R}^{256 \times 512}$).

\subsubsection{Decoder}
The decoder reconstructs the full-resolution field prediction through bilinear upsampling and skip connection fusion. At each scale, an UpSample block applies bilinear 2× upsampling followed by a 3×3 convolution and SiLU, then concatenates the corresponding encoder skip feature along the channel dimension, and processes the result through a FuseBlock consisting of a 3×3 projection convolution and a sequence of ResBlocks.

Sigmoid activations at all output heads bound predictions to [0, 1], which is required for numerical stability of the SSIM loss.

\subsection{Loss Function}
Training uses a multi-scale composite loss combining Structural Similarity Index Measure (SSIM) [8] and L1 losses at all three decoder output scales.
For the full-resolution (128×128) output:
\begin{equation}
    \mathcal{L}_{128} = \bigl(1 - \mathrm{SSIM}(\hat{\mathbf{y}}_{128}, \mathbf{y})\bigr)
    + \mathcal{L}_{L1}(\hat{\mathbf{y}}_{128}, \mathbf{y})
\end{equation}

For the auxiliary lower-resolution outputs, ground-truth fields are downsampled
via average pooling (factor 2 for $64\times64$, factor 4 for $32\times32$):

\begin{align}
    \mathcal{L}_{ms} =\; &0.5 \cdot \Bigl[\bigl(1 - \mathrm{SSIM}(\hat{\mathbf{y}}_{64},\, \mathbf{y}_{64})\bigr)
    + \mathcal{L}_{L1}(\hat{\mathbf{y}}_{64},\, \mathbf{y}_{64})\Bigr] \notag \\
    +\; &0.25 \cdot \Bigl[\bigl(1 - \mathrm{SSIM}(\hat{\mathbf{y}}_{32},\, \mathbf{y}_{32})\bigr)
    + \mathcal{L}_{L1}(\hat{\mathbf{y}}_{32},\, \mathbf{y}_{32})\Bigr]
\end{align}

The total loss is:
\begin{equation}
    \mathcal{L}_{\mathrm{total}} = \mathcal{L}_{128} + \mathcal{L}_{ms}
\end{equation} 

\section{Results and Discussions}
\begin{figure}[htbp]
\centerline{\includegraphics[width=0.5\textwidth]{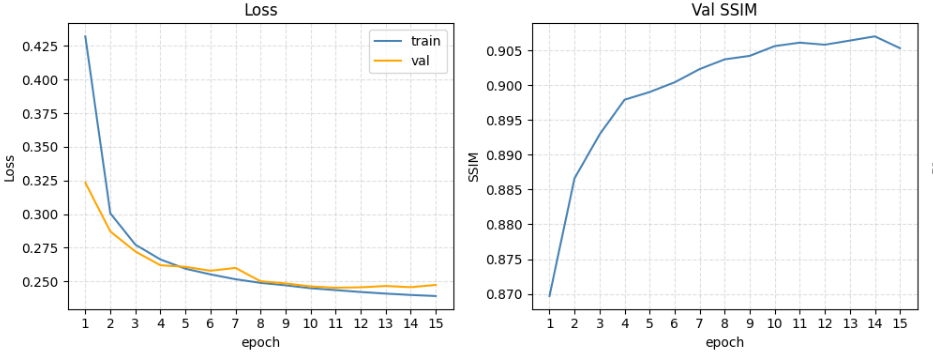}}
\caption{(a) Training Logs (b) Validation SSIM over training}
\label{fig:Training}
\end{figure}

Figure ~\ref{fig:Training} presents the training and validation loss curves
alongside the validation SSIM trajectory.
Validation SSIM improves
steadily from 0.8697 to 0.9070.

\subsection{Quantitative Performance}
Table~\ref{tab:results} summarizes the per-sample prediction metrics
across all eight subdatasets evaluated on the held-out test split.
The model is assessed using Mean Absolute Error (MAE), Structural
Similarity Index Measure (SSIM), and the coefficient of determination
$R^2$, computed between the predicted field and the corresponding FEM
ground truth.

\begin{table}[h]
\centering
\caption{Per-sample prediction metrics on the test split across all
subdatasets.}
\label{tab:results}
\begin{tabular}{lccc}
\hline
\textbf{Subdataset} & \textbf{MAE} & \textbf{SSIM} & $\mathbf{R^2}$ \\
\hline
BC\_scale               & 0.0050 & 0.9845 & 0.9991 \\
TRIANGLE                & 0.0052 & 0.9936 & 0.9980 \\
HEXAGON                 & 0.0129 & 0.9779 & 0.9935 \\
STITCH (GEO\_MISES)     & 0.0109 & 0.9730 & 0.9929 \\
PE12                    & 0.0266 & 0.8995 & 0.9617 \\
S12                     & 0.0445 & 0.8197 & 0.9240 \\
PE11\_0.2               & 0.0512 & 0.8543 & 0.8258 \\
S11                     & 0.0770 & 0.7126 & 0.7735 \\
\hline
\end{tabular}
\end{table}

The results reveal a consistent performance stratification across
subdatasets that correlates with the geometric regularity and field
complexity of each case. The boundary condition subdataset
(BC\_scale) and the TRIANGLE tessellation achieve the highest
accuracy of all evaluated cases, with $R^2$ values of 0.9991 and
0.9980 and MAE below 0.006 in both instances, indicating
near-perfect spatial and intensity agreement with the FEM reference.
The HEXAGON and STITCH subdatasets follow closely, with $R^2$ above
0.99 and SSIM above 0.97, confirming that the model generalizes
effectively to both periodic non-square tessellations and
high-resolution 32$\times$32 microstructural geometries.

PE12 and S12 remain quantitatively strong, with $R^2$
values of 0.9617 and 0.9240 respectively, though SSIM drops below
0.90 in both cases, reflecting the model's tendency to apply spatial
smoothing at sharp phase-boundary stress concentrations under
irregular block arrangements. PE11\_0.2 exhibits a further reduction
to $R^2 = 0.8258$ and SSIM $= 0.8543$, attributable to the sparse
geometry and extended low-stress brittle regions that reduce the
density of learnable structural cues available to the encoder.

\begin{figure}[htbp]
\centerline{\includegraphics[width=0.5\textwidth]{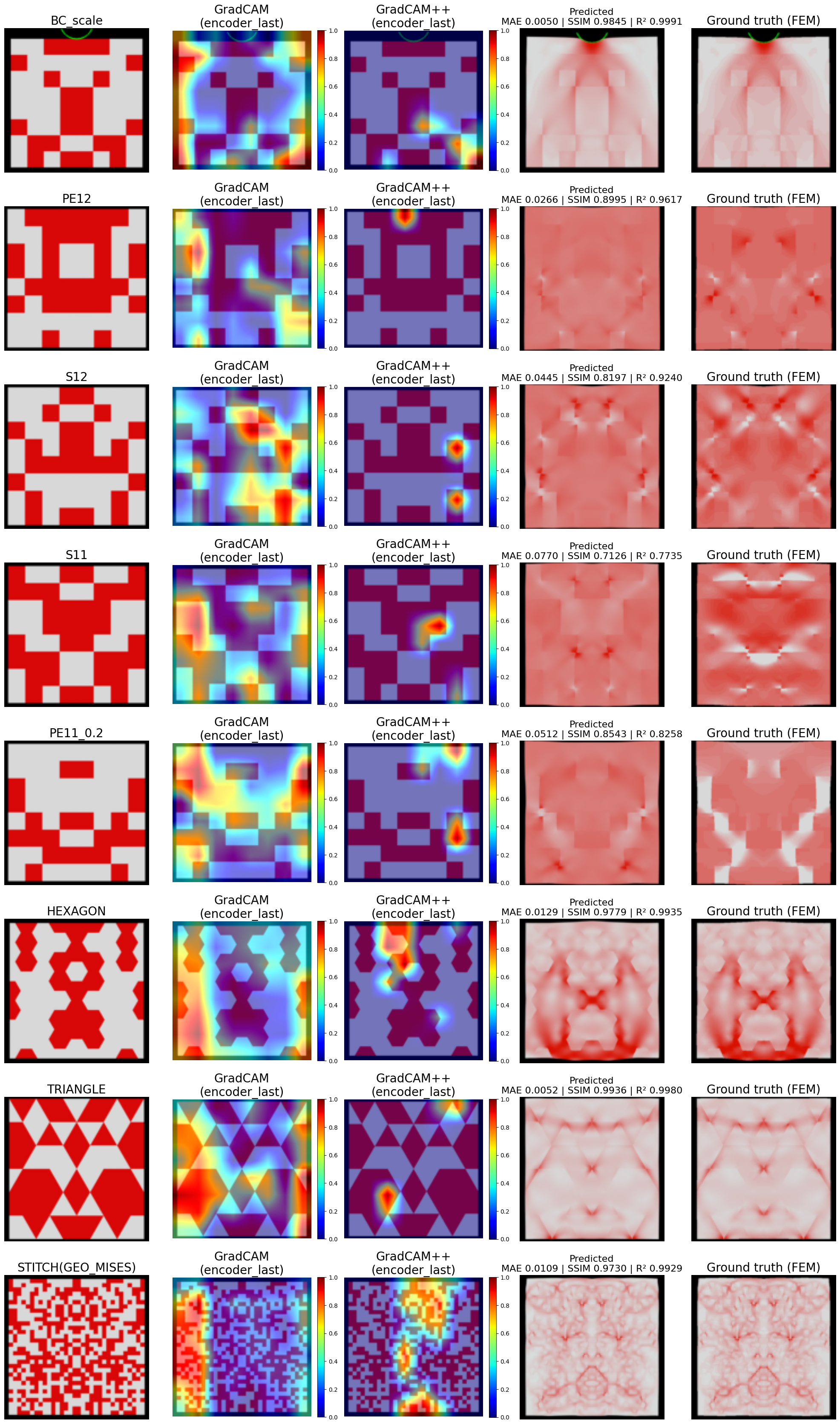}}
\caption{(a) Composite Input (b) GradCAM Results (c) GradCAM++ Results (d) Predicted Strain Fields (e) Ground Truth}
\label{fig:Results}
\end{figure}

\subsubsection{Qualitative Analysis}
Figure ~\ref{fig:Results} presents qualitative predictions alongside Gradient-weighted Class Activation Maps (GradCAM)[9] and GradCAM++[11] computed at the final encoder convolutional layer, for each of the three subdatasets.

\textbf{Boundary Condition Scale} The model achieves near-perfect performance (MAE 0.0050, SSIM 0.9845, R² 0.9991), accurately reproducing the arch-shaped stress diffusion beneath the spherical indenter. GradCAM shows broad activations originating at the indenter contact zone spreading downward, while GradCAM++ tightly concentrates at the top-center contact point  confirming the encoder has correctly parsed the green boundary condition marker as a mechanically meaningful spatial cue.

\textbf{PE12} The shear strain field is recovered with strong accuracy (MAE 0.0266, SSIM 0.8995, R² 0.9617). GradCAM exhibits a dominant activation band in the upper-left region extending diagonally toward center, consistent with the shear-dominant load path through this geometry. GradCAM++ sharpens this to a single hotspot at the primary block junction. Minor underestimation of peak shear concentrations at sharp corners accounts for the slight SSIM reduction.

\textbf{S12} Performance remains acceptable (MAE 0.0445, SSIM 0.8197, R² 0.9240) though SSIM drops relative to PE12, reflecting the more irregular block arrangement and sharper stress discontinuities. GradCAM activations are spatially scattered across the central region, while GradCAM++ identifies two discrete hotspots in the lower half corresponding to the most geometrically irregular phase junctions. The predicted field captures global topology but applies smoothing at the sharpest concentration sites.

\textbf{S11} This subdataset gave these results (MAE 0.0770, SSIM 0.7126, R² 0.7735). The FEM ground truth contains sharp, high-contrast stress discontinuities at phase boundaries that the model fails to reproduce at full intensity, producing an over-smoothed prediction. GradCAM shows diffuse, poorly localized activations with no dominant load path identified, and GradCAM++ provides only marginal improvement in spatial specificity  both indicating the encoder has not developed a geometrically precise internal representation for this field type.

\textbf{PE11-0.2} Performance is moderate (MAE 0.0512, SSIM 0.8543, R² 0.8258). The sparse geometry, dominated by large brittle regions with isolated soft inclusions, creates an irregular stress topology that the model partially recovers. GradCAM activations concentrate in the upper-center region while GradCAM++ identifies two hotspots in the lower-right quadrant  a spatial mismatch with the actual stress maxima location that directly explains the residual prediction error.

\textbf{HEXAGON} Strong performance (MAE 0.0129, SSIM 0.9779, R² 0.9935) reflects the model's effective exploitation of the periodic hexagonal geometry. GradCAM shows broad lateral activations along hexagonal boundary columns consistent with distributed load-sharing under compression. GradCAM++ localizes two symmetric hotspots at dominant ligament junctions, spatially aligned with FEM-computed stress maxima  confirming physically meaningful encoder representations for regular tessellation geometries.

\textbf{TRIANGLE} The highest combined accuracy in the figure (MAE 0.0052, SSIM 0.9936, R² 0.9980). GradCAM shows broad distributed activations across the full triangular lattice reflecting global load-sharing, while GradCAM++ sharpens to a single concentrated hotspot at the most mechanically loaded triangular junction in the upper-center. The tight spatial correspondence between GradCAM++ attention and FEM stress maxima confirms the encoder captures angular load-transfer geometry rather than superficial color patterns.

\textbf{GEO-MISES} The high-resolution 32×32 microstructure presents the greatest geometric complexity in the figure. Despite this, the model achieves strong performance (MAE 0.0109, SSIM 0.9730, R² 0.9929), recovering the branched stress propagation patterns visible in the FEM ground truth. GradCAM activations are broadly distributed reflecting the high information density of the fine-grained geometry, while GradCAM++ identifies a concentrated central vertical band corresponding to the dominant vertical stress transmission path through the connected phase. The slight MAE increase relative to simpler geometries reflects the higher spatial complexity rather than a model generalization failure.

Across all subdatasets, GradCAM consistently highlights mechanically meaningful regions  structural junctions, load-bearing ligaments, and material boundaries  rather than geometrically trivial areas such as void interiors or uniform-field zones. On the other side, GradCAM++, highlights the concentrated regions which have dominant stress or strain fields. This confirms that the Transformer bottleneck and Unet layers can together learn spatially coherent, physically grounded feature representations directly from image supervision.

\section{Conclusion and Future Works}
This paper presented, a hybrid Unet-Transformer surrogate model for direct prediction of FEA stress and strain fields from geometry images. The proposed architecture combines a four-stage residual convolutional encoder, a six-layer Transformer bottleneck with 2D sinusoidal positional embeddings, and a multi-scale decoder with skip connections, trained end-to-end using a composite SSIM and L1 loss across ten FEA subdatasets simultaneously. Experimental evaluation on the HEXAGON, PE12, and S12 subdatasets demonstrates that the model achieves strong quantitative accuracy  with $R^2$ values above 0.96 and MAE below 0.03 across all subdatasets  while GradCAM analysis confirms that the learned representations align with physically meaningful stress-concentrating regions, without any explicit mechanical supervision. The results establish that a single unified model can generalize across structurally diverse geometry types and field quantities within the same training framework.

These findings suggest that hybrid Unet-Transformer architectures are a viable and efficient alternative to iterative FEM solvers for engineering design exploration, offering near-instant field predictions at inference time once trained. For the future works, the architecture can be evolved to for 3D Unet with transformer as bottleneck for predicting 3D stress and strain directions in composite material.

\end{document}